\documentclass{article} 

\usepackage{nips11submit_e,times}

\usepackage{graphicx} 
\usepackage{subfigure}

\usepackage[numbers]{natbib}

\usepackage[ruled]{algorithm2e}

\usepackage{hyperref}
\usepackage{amsmath}
\usepackage{amsfonts}
\usepackage{amssymb}
\usepackage[normalsections,normalmargins,normalindent,normalleading,normaltitle,normalbib]{savetrees}

\title{Learning Representations by Maximizing Compression.}

\author{
Karol Gregor and Yann LeCun\\
Department of Computer Science\\
Courant Institute, NYU\\
715 Broadway Floor 12\\
New York, NY, 10003\\
\texttt{karol.gregor@gmail.com,yann@cs.nyu.edu}\\
}

\nipsfinalcopy 

\begin{document}

\maketitle
\begin{abstract}
We give an algorithm that learns a representation of data through compression. The algorithm 1) predicts bits sequentially from those previously seen and 2) has a structure and a number of computations similar to an autoencoder. The likelihood under the model can be calculated exactly, and arithmetic coding can be used directly for compression. When training on digits the algorithm learns filters similar to those of restricted boltzman machines and denoising autoencoders. Independent samples can be drawn from the model by a single sweep through the pixels. The algorithm has a good compression performance when compared to other methods that work under random ordering of pixels.
\end{abstract}

\section{Introduction}


Modeling data essentially means the ability to predict new data from previously seen data. It is generally impossible to predict data exactly and at best one can assign probabilities to various outcomes. Modeling nonsequential data such as images can also be viewed in this framework - as predicting images from no other data. Difficulties in modeling data lie in the fact that in high dimensions the distribution can be complex,  multimodal, and it is hard to train models to maximize data likelihood. Special systems have been devised where this is tractable, or where other reasonable criterion is optimized. These include restricted bolzman machines (RBM) \citep{hinton2006fast}, score matching \citep{hyvarinen2006estimation}, herding \citep{welling2009herding}, denoising autoencoders \citep{vincent2008extracting} or sparse coding \citep{olshausen1996emergence}.

\begin{figure}[h!]
\begin{center}
\includegraphics[width=0.85\textwidth]{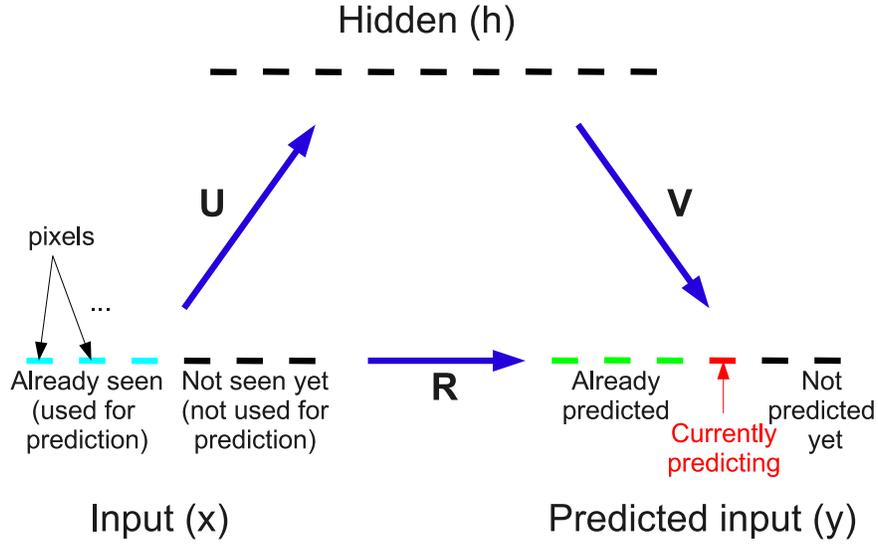}
\end{center}
\caption{A diagram of our system. The system predicts pixels of the input sequentially. A given pixel is predicted only from pixels already seen. The prediction function consists of two parts. The $R$ path is a direct matrix multiplication. The $UV$ path is a matrix multiplication followed by sigmoid and followed by a matrix mupltiplication. The two paths are added and passed through final sigmoid to generate the prediction of the next pixel.}
\label{fig_prediction_figure}
\end{figure}

In this paper we approach data modeling as follows. We observe that, unlike high dimensional distributions, predicting a single bit is easy and is completely specified by a single number - the probability of that bit being one. To use this observation we put all bits (e.g. pixels of a binary image) into a sequence and write a model that predicts the next bit from the bits already seen. We give the model the structure of an autoencoder as follows. We connect the input to a hidden layer and the hidden layer to the predicted input, both through a matrix and sigmoid nonlinearity, Figure \ref{fig_prediction_figure}. When a new pixel arrives the values in the hidden layer change by the product of the pixel value and the strength of the connectinons between the pixel and the hidden layer. The new values in the hidden layer are then used to predict the value of the next pixel. We also include direct input to predicted input connections. 


Predicting new pixel from previous ones was done before \citep{bottou1998high,graves2008novel}. The former is similar to our system without hidden layer but assumes spatial layeout of pixels. The latter uses a model that takes previous bits (and possibly values of a hidden layer) and calculates the full new value of the hidden layer. In our case we are building up one model of the input, and when new pixel arrives we only use that pixel to update the value of the hidden layer. The model has the struncture of an autoencoder and the final representation (after training) can be obtained directly by matrix muplitplication followed by sigmoid nonlinearity.

We train our model on USPS 
and MNIST \citep{lecun1998gradient} digits. We find that learned filters learn strokes of digits and in appropriate setting are similar to those of RBM's or denosing autoencoders. The model can generate independent random samples of digits by sweeping accross all the pixels each time predicting new pixel, sampling and using that value as known value of the pixel at the next iteration. The model generates reasonable images of digits. Since the model generates probability of the next bit, arithmetic coding can be used directly for compression and the likelihood under the model is the encoded number of bits. The algorithm outperforms other related methods on compression.

\section{The model}

We assume that the input is a binary vector $\mathbf{x}$. We pick a random ordering of its pixels (bits) $x_1,x_2,\ldots,x_{n_x}$ and build a model $P(x_k|x_{k-1},\ldots,x_1)$. The probability of a given image under this model is then $P(x_1,\ldots,x_{n_x})=P(x_{n_x}|x_{n_x-1},\ldots,x_1)\cdots P(x_1)$. We define the model as follows, Figure \ref{fig_prediction_figure}. There is an input layer $\mathbf{x}$, predicted input layer $\mathbf{y}$ and a hidden layer $\mathbf{h}$. The following are the parameters of the model: matrix $\mathbf{U}$ between input and hidden layer, matrix $\mathbf{V}$ between the hidden and predicted input layer, matrix $\mathbf{R}$ between the input and predicted input layer, bias vector $\mathbf{b}_h$ in the hidden layer and bias vector $\mathbf{b}_y$ in the predicted input layer. Let $\bar{x}=x-x_{ave}$ where $x_{ave}$ is the average (image) over all training images. Let $\bar{\mathbf{x}}^k$ denotes an input vector where $\bar{x}^k_i=\bar{x}_i$ for $i \leq k$ and $0$ otherwise. We iterate over the pixels from the beginning till the end. At $k$-th iteration we have seen pixels $1,\ldots,k$ and our goal is to predict the $k+1$-th pixel. The value of the hidden layer, and the predicted value of the $k+1$-th pixel are
\begin{eqnarray}
\label{eq_h}
\mathbf{h} &=& \mathbf{\sigma} (\mathbf{U} \cdot \mathbf{\bar{x}}^k + \mathbf{b_h}) \\
y_{k+1} &=& \sigma (\mathbf{V}_{k+1} \cdot \mathbf{h} + \mathbf{R}_{k+1} \cdot \bar{\mathbf{x}}^k + b_{y,k+1}) 
\label{eq_x}
\end{eqnarray}
where $\mathbf{V}_{k+1}$ is the $k+1$-th row of $\mathbf{V}$. The bold letters denote vectors or matrices and others denote numbers. The goal is to maximize the sum of the log likelihoods of these predictions (which equals the log likelihood of the input) which equals
\begin{equation}
L=-\sum_i x_i \log_2(y_i) + (1-x_i) \log_2 (1-y_i)
\label{eq-cost}
\end{equation}

Let us look at the number of calculations involved. Given a new pixel $k$ to update the hidden layer (\ref{eq_h}) costs $n_h$ operations because the previous $\mathbf{U} \cdot \bar{\mathbf{x}}^{k-1}$ have already been calculated. The $\mathbf{V}_{k+1} \cdot \mathbf{h}$ term in (\ref{eq_x}) costs $n_h$ operations and the $\mathbf{R} \cdot \bar{\mathbf{x}}^k$ term costs $n_x$ operations because $\mathbf{R} \cdot \bar{\mathbf{x}}^{k-1}$ has already been calculated. The $\mathbf{b}_h$ and $\sigma$ in (\ref{eq_h}) costs $n_h$ operations and the $b_{y,k+1}$ and $\sigma$ in (\ref{eq_x}) costs $1$ operation. Thus the total number of operations is $O(n_h n_x+n_x n_x)$. In the absence of the $\mathbf{R}$ term the actual number of operations is the same as that of an autoencoder, except that the $\mathbf{b}_h$ and $\sigma$ in (\ref{eq_h}) has to be evaluated $n_x$ times instead of once. This could be a problem since the nonlinearity is costly. We precompute it and the resulting time is comparable to the other operations. Thus this algorithm (without $\mathbf{R}$) has similar number of calculations as autoencoder. The $\mathbf{R}$ doesn't need to be present, but it improves the performance slightly. 

The training is done by backpropagation. In the next section we write the full algorithm and see that the backpropagation has a similar number of computations as the forward part.

\begin{algorithm}[H]
$\mathbf{x^k}=0$\\
$x_0=1$\\
$\mathbf{U_0} \equiv b_h$\\
$\mathbf{V}_{n+1}=\mathbf{0}$\\
$\mathbf{y^v}=\mathbf{b_y}$\\
\For{$j=0:n$}{
  $\mathbf{h}^u=\mathbf{h}^u+\mathbf{U}_j \bar{x}_j$ \\
  $\mathbf{h}=\sigma(\mathbf{h}^u)$ \\
  $\mathbf{h^{save}}_j=\mathbf{h}$\\
  $x^k_j=\bar{x}_j$\\
  $y^v_{j+1}=y^v_{j+1}+\mathbf{V}_{j+1} \cdot \mathbf{h} + \mathbf{R}_{j+1} \cdot \mathbf{x}^k$ \\
  The prediction and the number of bits needed to encode this pixel\\
  $y_{j+1}=\sigma(y^v_{j+1})$ \\
  $L_{j+1}=-x_{j+1}\log_2(y_{j+1})-(1-x_{j+1})\log_2(1-y_{j+1})$
}
$d\mathbf{y}=-\frac{\mathbf{x}}{\mathbf{y}}+\frac{1-\mathbf{x}}{1-\mathbf{y}}$\\
$d\mathbf{y}^v=d\mathbf{y}(\mathbf{y}-\mathbf{y}^2)$\\
$\mathbf{b_y}=\mathbf{b_y}-\eta d\mathbf{y}^v$\\
\For{$j=n:-1:0$}{
  $\mathbf{h}=\mathbf{h}^{save}_j$\\
  $x^k_{j+1}=0$\\
  $\mathbf{R}_{j+1}=\mathbf{R}_{j+1}-\eta dy^v_{j+1} \mathbf{x}^k$\\
  $\mathbf{V}_{j+1}=\mathbf{V}_{j+1}-\eta dy^v_{j+1} \mathbf{h}$\\
  $d\mathbf{h}=\mathbf{V}_{j+1} dy^v_{j+1}$\\
  $d\mathbf{h^v}=d\mathbf{h}(\mathbf{h}-\mathbf{h}^2)$\\
  $\mathbf{U}_j=\mathbf{U}_j-\eta d\mathbf{h}^v \bar{x}_j$\\
}

\caption{Sequential Pixel Prediction}
\end{algorithm}

\section{The Algorithm}

The full algorithm is displayed in Algorithm 1. Its first part (up to including the first for loop) was outlined in the previous section and the second part is the backpropagation of the first part. The $\eta$ is the learning rate. The $y-y^2$ and $h-h^2$ terms come from derivative of sigmoid (if $y=\sigma(x)$ then $dy/dx=y-y^2$). The $\mathbf{h}^{save}$ are the saved values of each $\mathbf{h}$ at each iteration. This is needed for backpropagation. Alternativelly, to save memory at the expense of computation cost, one can obtain this value by progressivelly subtracting $\mathbf{U}_j\bar{x_j}$ from $\mathbf{h}^v$. As explained in the previous section the total computational cost consist of costs of: matrix multiplications by $\mathbf{U}$, $\mathbf{V}$, $\mathbf{R}$ and $n_x$ evaluations of the $\sigma$. The last step is rather costly if one uses built in functions but comparable to the $\mathbf{U}$ matrix multiplication if one uses precomputed sigmoid. The cost of backpropagation is similar to that of the forward part. 

\section{Training}

We use USPS and MNIST digits as our datasets. The pixel values go between 0 and 255. We binarize these datasets by setting new pixel value to $0$ if the old one is below a theshold and $1$ if it is above. We use threshold of $50$ for USPS and $128$ for MNIST. 

We train the system using stochastic gradient descent: We loop over the Algorithm 1 each time picking a random digit from the training set. The learning rate $\eta$ decays as $\sim 1/t$. 

We choose the pixel permutation in the following ways: 1) Different random permutation at each iteration 2) Fixed random permutation 3) Going from upper left to lower right as if reading the page. We found the 1) leads to ``nicer looking'' filters then 2) and 3) but leads to worse performance on compression. The 2) and 3) have similar performance on compression.

\section{Learned Filters}

\begin{figure}[h!]
\begin{center}
\includegraphics[width=.92\textwidth]{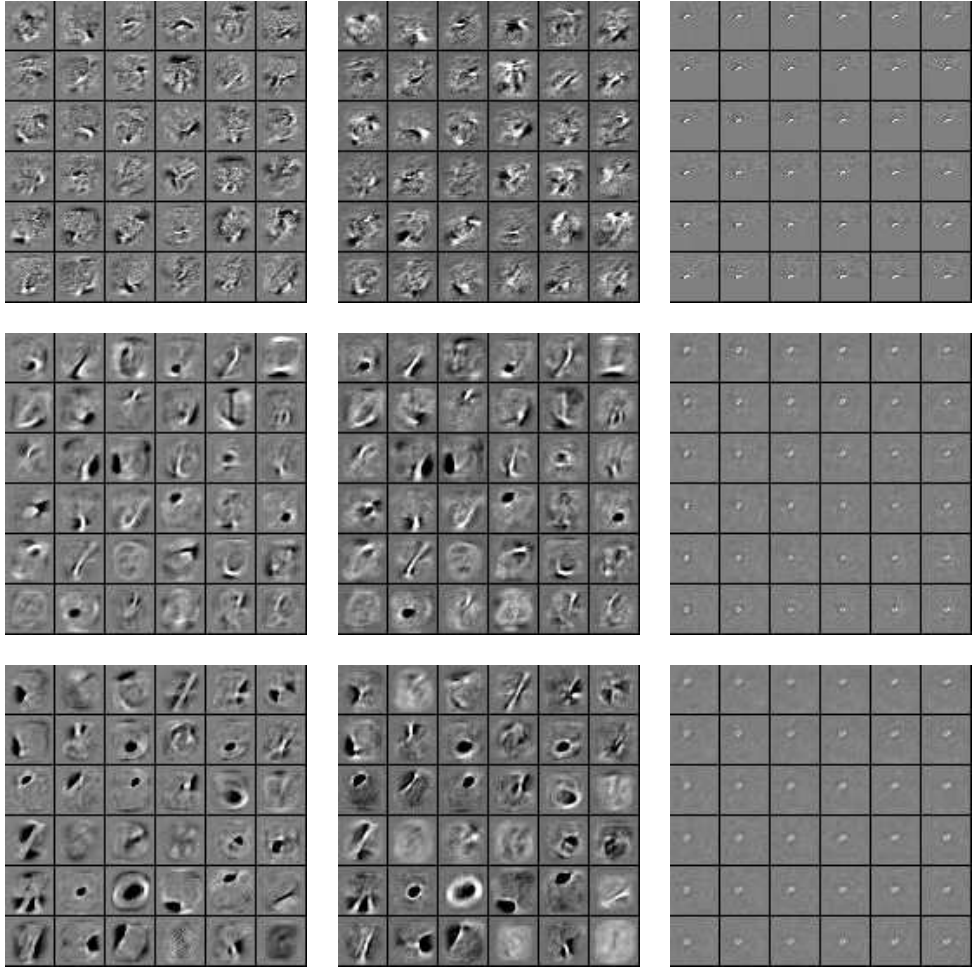}
\end{center}
\caption{Examples of filters learned on MNIST digits. The first column is the U matrix, the second column is the V matrix and the third column is the R matrix. In the first row the permutation of pixels went from upper left to lower bottom (reading page). In the second row the permutation was random and changed at each iteration. In the third row the permutation was the same as in the previous row, but in addition the average pixel values were not subtracted from the input ($\bar{x}=x$).}
\label{fig_filters}
\end{figure}

Learned filters are displaed in the Figure \ref{fig_filters}. We see that random permutation at each iteration Figure \ref{fig_filters}b,c results in filters containing strokes or localied circular features. The when no subtraction by average digit is used in the input, the filters resemble more those of RBM's or denoising autoencoders.

\section{Generated data}

We generate data from the model as follows. We put the pixels in the same sequence as in training. We sample the first pixel as if all input pixels were zero (from binary distribution). Given a number of sampled pixels we calculate probability of the next pixel from already sampled pixels using the model and then sample.

\begin{figure}[h!]
\begin{center}
\includegraphics[width=0.99\textwidth]{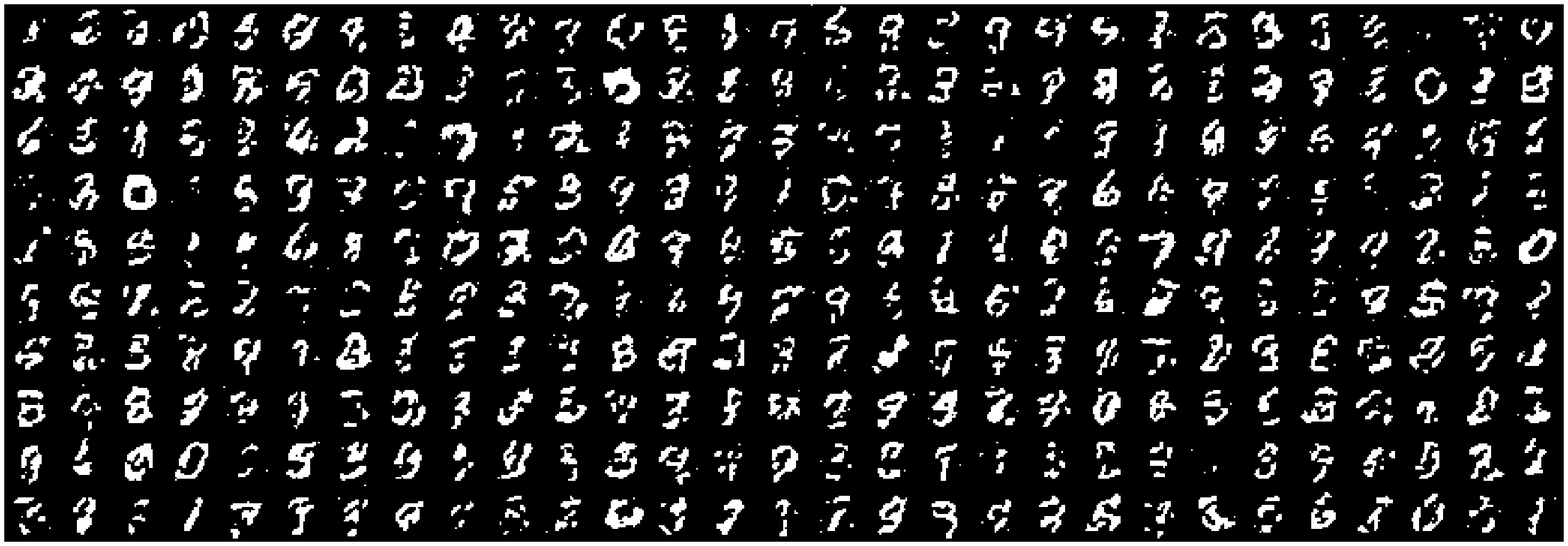}
\end{center}
\vspace{0.1cm}
\begin{center}
\includegraphics[width=0.99\textwidth]{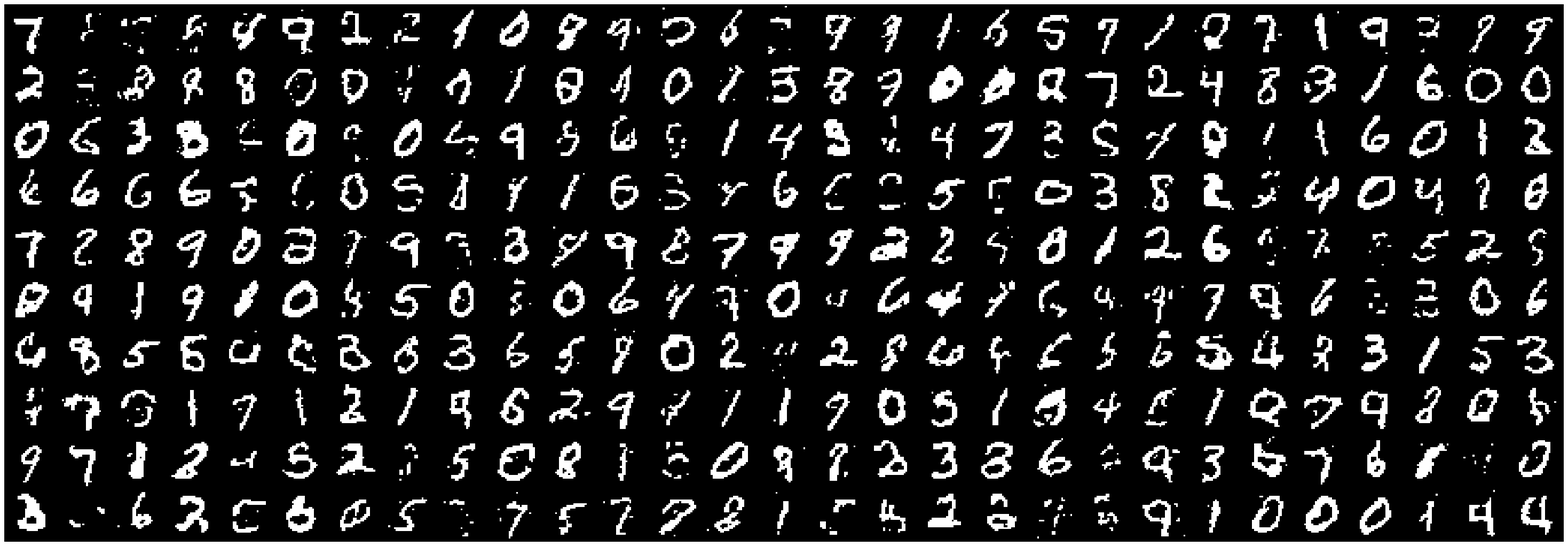}
\end{center}
\caption{Generated digits after training on MNIST with a) the system with R path only b) the full system. For each image the pixels were generated in sequence from upper left to lower right. Given $n$ already generated pixels, the new pixel was generated by calculating its probability under the model and sampling. The a) mostly captures the local structure but b) captures full structure and often generates nice digit. Note that this way we obtain independent samples, each containing one sweep throug the pixels.}
\label{fig_gen_dig}
\end{figure}

The generated samples under the full model are shown in the Figure \ref{fig_gen_dig}. We see that many of these are nice digits.

\section{Compression}

We calculate compression in the following framework. Imagine we want to transmit information to another other person. We can agree ahead on the model that we use (as in jpg for example). That model can be trained on some data. Now given this model we want to transfer new data and calculate compression on them. In this paper we model this by training the system on the training set and calculating the compression on the testing set (new unseen data). For MNIST we use the split already provided and for USPS we randomly choose 700 images per class for training and 300 images per class for testing.

\subsection{Benchmark 1: Difference from the nearest neighbor center}

For compression using arithmetic coding one needs a probability $p$ for a given pixel at the time of the encoding. The resulting number of bits equals the sum of $x\log_2 (p)+(1-x)\log_2 (1-p)$ over all the data (pixels of all images) where $x$ is the actuall value of the pixel.

The original data, where one pixel needs one bit, corresponds to setting $p$ to $0.5$ for all pixels. This gives $16 \times 16 = 256$ bits for USPS and $28 \times 28 = 784$ bits for MNIST per digit. A better encoding is to use a constant value of $p$ different from $0.5$. It is easy to show that to maximize compression on the training set, one needs to use $p$ that equals the average value of all pixels over training set. On the test set this results in encoding of $236.5$ bits on USPS and $441.8$ bits on MNIST. Even better compression is obtained by using different value of $p$ for each pixel (but the same for a given pixel accross images). This results in encoding of $219$ bits for USPS and $297$ bits on MNIST. 

As our first benchmark we use a better compression scheme. We pick at random a given number of inputs from the training set and call them centers. In order to encode a given (new) input we choose the center closest to the input and encode the difference of this input from the center. To encode the difference we again calculate the probability $p$ that a given pixel of the given input is different from the center and use this value in arithmetic coding. There will be different value for each pixel and each center. The total number of bits needed to encode a given input consist of number of bits needed to specify the which center is the closest ($\log_2 N_{centers}$) and the number of bits needed to specify the difference.

There is one issue that needs to be addressed before this can work. It happens that there are some pixels of a given center for which all the inputs in the training set had the same value at that pixel. Consequently the $p$ for that pixel would be $0$ or $1$. However in the test set it can happen that the pixel value at that location is different from the traing set value and then to encode that pixel we would need $-\log_2 0=\infty$ bits. To remedy this problem we regularize by bounding $p$ to be between $\epsilon$ and $1-\epsilon$. We crossvalidate over the $\epsilon$ to find the regularization that leads to the best compression. On USPS dataset this leads to encoding of digits with $140$ bits and happens with approximatelly $1000$ centers and on MNIST the best encoding has $178$ bits and happens at approximatelly $2000$ centers.

\subsection{Benchmark 2: DjVu}

We use the basic compression algorithm of DjVu \citep{bottou1998high} which is as follows. One goes over pixels from upper left to lower right as if reading the page and as in our system, predicts the current pixel from previous ones. One chooses ten previous pixels that are located on the left or above the current pixel, see \citep{bottou1998high}. The model used to predict the current pixel is the following. There are $1024$ possibilities for the values of the ten pixels just mentioned. One loops over the training set and locations. Each such instance belongs to one of the $1024$ possibilities and sometimes leads to the value of the new pixel being one and sometimes zero. The average of these outcomes is the predicted probability $p$ of the value of the pixel.

Note that this algorithm assumes that specific layout of pixels on two dimensional grid (image) whereas our algorithm doesn't and pixels can be permuted. On the other hand it should also be mentioned that this is the basic version and in the actual DjVu a different model is used based on context. It is possible that this would improve the performance. However in our case one can also imagine making our model context dependent, and it is left for future work.

\subsection{Compression results}

We train the system as explained above. The resulting performance on the training set is better then on the test set. Without regularization this is increasingly true as we increase the number of hidden units. For example without regularization, the training set performance with $1000$ units on USPS is $47.1$ bits but the testing set performance is $83.6$ bits. To remedy this we regularize with $L_2$ norm on all the parameters. We also observe that the testing error goes down at first and then starts increasing. For example in the example just mentioned, it increases from the $83.6$ bits to $92$ bits. We therefore use early stopping.

\begin{table}
\caption{Comparison of various methods on digit compression}
\begin{center}
\begin{tabular}{|l||l|l|}
\hline
 &\multicolumn{2}{l|}{Encoding (bits)}\\
\cline{2-3}
Method & USPS & MNIST\\
\hline\hline
Original Image & 256 & 784\\
\hline
Pixel independent probability & 236 & 442 \\
Pixel dependent probability & 219 & 297 \\
\hline
RBM \citep{chen2010parametric} & 149 & - \\
Herding \citep{chen2010parametric} & 144 & - \\
\hline
Difference from NN center & 140 & 178 \\
\hline
Context independent DjVu & 91.7 & 119\\
\hline
Spatial prediction, R only & 91.5 & 109 \\
Spatial prediction, 200 units, UV only & 87.4 & 97.8\\
Spatial prediction, 200 units, full & 84.5 & 94.8 \\
Spatial prediction, 400 units, full & 83.8 & 91.2 \\
Spatial prediction, 1000 units, full & 81.0 & 92.2 \\
\hline
\end{tabular}
\end{center}
\label{table_comp}
\end{table}

The results are displayed in the Table \ref{table_comp}. We see that the performance of DjVu and $R$ only system on USPS are essentially the same. This is reasonable since the learned filters of $R$ are very local and probably capture the same information as the local setting of DjVu. On MNIST the results are somewhat better for the $R$ system, probably because the digits are larger and one need somewhat larger region to predict the next pixel. Note that unlike DjVu, our method works under a random permutation of pixels.

The performance of this system is significantly better then that of the first benchmark and the other systems mentioned from literature. However the failing of RBM and Herding very likely does not lie in a bad representation but rather in the compression scheme. The compression scheme is more similar to the difference from the nearest neighbor center and the bad performance likely comes from difficulty of estimating the likelihood and the absence of natural compression scheme.

Finally we make a few observations. First, after our system is trained, the hidden layer given the input can be obtained in the same way as in RBM or an autoencoder - matrix multiplication (+ bias) followed by sigmoid. Second we binarized the pixel values in order to calculate the compression. If we are simply interested in hidden representation, we can keep the original (quasi-binary) values as is customary and use exactly the same algorithm. Third, we can chose any other nonlinearity in the hidden layer. Fouth, we can adjust algorithm for real valued data. Then, specifying distrinution over pixel means to specify distribution in one real variable. While that can potentially be complex, it is likely that gaussian would suffice in most circumstances. The comlexity and multimodality of the distribuition is obtained by composing simple one dimensional distriutions and sampling. 

\section{Conclusions}

We have intorduced a conceptually simple algorithm for representation learning. We showed that it performs very well on compression and that the learned representations are sensible. We can train the model efficiently and calculate likelihood of data under the model exactly. In the future our goal is to focus on building more complex context dependent models with several layers of representation.

\section{Acknowledgements}
We thank Arthur Szlam for useful discussions and for suggesting the first Benchmark. This work was funded in part by ONR project "Learning Hierarchical Models for Information Integration", and NSF EFRI-0835878: EFRI/COPN "Deep Learning in the Mammalian Visual Cortex"

\bibliographystyle{alpha}
\bibliography{spatialpred}

\end{document}